\documentclass[journal]{IEEEtran}
\ifCLASSINFOpdf
\usepackage[pdftex]{graphicx}
\else
\fi
\usepackage{cite}

\begin{document}
%
\title{Agentic AI-Empowered Conversational Embodied Intelligence Networks in 6G}
%
%
%

\author{Mingkai~Chen,~\IEEEmembership{Member,~IEEE, } Zijie~Feng,~\IEEEmembership{Student~Member,~IEEE, }
Lei Wang,~~\IEEEmembership{Member,~IEEE,} \\
Yaser Khamayseh,~\IEEEmembership{Member,~IEEE}
\thanks{M. Chen, Z. Feng, and L. Wang are with the Key Laboratory of Broadband Wireless Communication and Sensor Network Technology, Nanjing University of Posts and Telecommunications.  

Y. Khamayseh is with College of Technological Innovation
Zayed University
Abu Dhabi, United Arab Emirates.

Corresponding Authors are Lei Wang, wanglei@njupt.edu.cn}
}

\maketitle

\begin{abstract}
In the 6G era, semantic collaboration among multiple embodied intelligent devices (MEIDs) is becoming a key capability for complex task execution. However, existing systems remain some challenges on multimodal information fusion, adaptive communication, and decision interpretability, enabling efficient collaboration in dynamic environment. To address this, we propose a Collaborative Conversational Embodied Intelligence Network (CC-EIN) framework that integrates multimodal feature fusion, adaptive semantic communication, task coordination, and interpretability. First, a cross-modal fusion maps image and radar data into a unified semantic representation, ensuring consistent task understanding across MEIDs. Second, an adaptive semantic communication strategy dynamically adjusts coding schemes, compression ratios, and transmission power according to the urgency of the task and the channel conditions, thus improving spectrum efficiency under bandwidth constraints. Third, a semantic-driven collaboration mechanism decomposes and allocates tasks through a shared knowledge base, enabling drones, autonomous vehicles, and robot dogs to cooperate effectively while avoiding conflicts. Finally, decision visualization using Gradient-weighted Class Activation Mapping (Grad-CAM) highlights agents’ focus areas during decision-making, enhancing transparency and trust. Simulations show that the proposed framework achieves a 95.4\% task completion rate (TCR) and 95\% transmission efficiency (TE) in post-earthquake rescue scenarios, while showing significant advantages in semantic consistency (SC) and energy-adaptive performance.
\end{abstract}


%
\IEEEpeerreviewmaketitle

\section{Introduction}
With the gradual deployment and evolution of 5G, both academia and industry have increasingly focused on the development of 6G mobile networks. 6G is envisioned as the infrastructure for future intelligent societies, with the aim of providing ultra-high bandwidth, ultra-low latency, ubiquitous coverage, and advanced intelligence. According to \cite{1},  6G can be summarized with four key concepts: intelligent connectivity, deep connectivity, holographic connectivity, and ubiquitous connectivity, to meet the increased communication demands expected around 2030. Moreover, 6G is expected to deeply embed artificial intelligence into network systems, endowing the core network, edge, and end devices with distributed intelligence for end-to-end smart operation \cite{2}. In this context, future intelligent communication networks will evolve beyond simple connectivity, becoming intelligent systems with environmental awareness, autonomous decision making, and adaptive optimization, supporting emerging applications such as the industrial Internet, smart transportation, and unmanned systems \cite{3}.

With the rapid advancement of embodied intelligent device (EID), including robots, drones, autonomous vehicles, and humanoid robots, future intelligent systems will increasingly depend on their collaborative capabilities \cite{4}. Embodied intelligence highlights that agents interact dynamically with their environment through physical 'bodies', executing autonomous behavior via a perception–understanding–decision–action loop. These devices, equipped with diverse sensors and intelligent algorithms, can collect data, perceive the environment in real time, and make informed decisions. For instance, drones leverage visual sensors for precise navigation, autonomous vehicles use radar and LiDAR point clouds for environmental perception, and robot dogs perform tasks in complex terrains, such as post-disaster rescue and environmental exploration. Their high autonomy is driven by Agentic AI, an intelligent agent capable of autonomously perceiving, understanding, learning, and making decisions independently or collaboratively according to task requirements \cite{5}. As MEIDs become more widespread, Agentic AI-based intelligent agents are expected to be a key driver of future intelligent societies, reshaping operations across various industries.

Although individual EIDs can perform independent tasks, their capabilities are inherently limited in complex and dynamic scenarios \cite{6}. To overcome these constraints, the concept of Embodied Intelligence Networks (EINs) has emerged, where multiple embodied devices are interconnected to form a cooperative network. Such a network leverages distributed sensing, communication, and decision-making to accomplish tasks more efficiently and accurately than a single device. By enabling resource sharing and dynamic interaction, EINs exhibit collective behaviors that significantly enhance resilience and adaptability in disaster response and other challenging environments. However, as EINs become more prevalent, effective human–machine collaboration also becomes a critical issue \cite{7}. Users not only need to work alongside embodied devices but also understand their decision-making logic. Thus, interpretability of EINs decision-making is essential: transparent decision logic improves trust, strengthens emergency response capabilities, and supports rapid, coordinated actions in complex scenarios, ultimately facilitating smoother human–AI collaboration \cite{8}.

In an EIN, communication resources and dynamic network conditions pose significant challenges \cite{9}. Devices typically rely on self-organizing networks, which often suffer from limited bandwidth, frequent disruptions, and latency, making traditional static communication strategies insufficient for timely task information delivery. Moreover, when operating as part of an EIN, EIDs generate massive multimodal data streams—such as visual, audio, and LiDAR point clouds—further increasing the transmission burden. Traditional communication methods struggle to efficiently process and deliver such heterogeneous data. Semantic communication provides an effective solution: unlike conventional bit-level transmission, it adopts a task-oriented 'understand first, transmit later' paradigm \cite{10}. Task-relevant semantic features are extracted from raw multimodal data, compressed, and transmitted, before being reconstructed at the receiver for downstream task execution. This approach not only reduces transmission overhead and improves efficiency under bandwidth constraints but also enables shared semantic understanding within the EIN. By leveraging semantic information, devices within the EIN can engage in collaborative reasoning and task allocation, thereby overcoming the limitations of traditional methods in dynamic environments and ensuring that complex tasks are completed reliably \cite{11}.

In response to this situation, this article proposes the CC-EIN framework, which integrates four core components: perception, communication, collaboration, and interpretability to systematically support efficient MEIDs coordination. The main contributions of this work are as follows.
\begin{itemize}
\item We propose the Perception \& Semantic Network (PerceptiNet), which integrates multimodal perception data to achieve comprehensive environmental understanding. It performs a deep fusion of multimodal data and extracts key semantic features to offer an environmental perception.
\item The Dynamic Resource Allocation Optimization for Semantic Communication (DRAOSC) network is designed to adjust semantic transmission strategies based on task urgency, communication conditions, and network status, allowing adaptive resource allocation and optimization. It ensures efficient and reliable delivery of key information while avoiding bandwidth waste and latency issues typical of traditional communication methods, significantly improving network adaptability and efficiency.
\item The Cohesive Multi-agent Decision Making (CohesiveMind) is designed to provide MEIDs with intelligent task collaboration capabilities. It decomposes tasks into semantic information, optimizes resource management, and facilitates efficient collaboration among devices. By minimizing conflicts and redundant operations, it ensures smooth and successful task execution.
\item The Interpretability of Decision (InDec) is introduced to provide a transparent decision-making process for the system. It visualizes the decision processes of EIDs, allowing operators to clearly understand the rationale behind the device decisions. It enhances the transparency and accountability while improving the efficiency and trustworthiness of human–machine collaboration.
\end{itemize}

The remainder of this article is organized as follows. Section \uppercase\expandafter{\romannumeral2} reviews related work and highlights the outstanding challenges. Section \uppercase\expandafter{\romannumeral3} introduces the proposed framework, which includes multimodal information fusion, adaptive semantic communication, task collaboration mechanism, and interpretability. Section \uppercase\expandafter{\romannumeral4} presents framework performance evaluations in post-disaster rescue scenarios. Finally, Section \uppercase\expandafter{\romannumeral5} concludes the article and discusses directions for future research.

\section{Open Issues in Embodied Intelligence Network }

\subsection{Overview}
The EINs treat MEIDs as cooperative units, achieving joint perception, decision-making, and execution through shared semantic representations and coordination strategies \cite{12}. Instead of exchanging raw data, devices compress and transmit information based on task semantics, enabling cross-device task decomposition and scheduling. EINs adopt an adaptive communication mechanism that dynamically adjusts encoding, scheduling, and power according to task urgency and network conditions, ensuring reliable transmission of critical semantics under bandwidth limitations or link fluctuations. By tightly coupling cooperation with adaptive communication, EINs enable faster response and higher resource efficiency in complex scenarios such as disaster rescue and intelligent transportation \cite{13}. However, it still faces challenges in perception, communication and decision-making, including multimodal information integration and dynamic resource allocation, while ensuring reliability and interpretability. Therefore, we list some potential challenges as follows.

\subsection{Challenges}
\begin{figure}[htbp]
\centering
\includegraphics[scale=0.24]{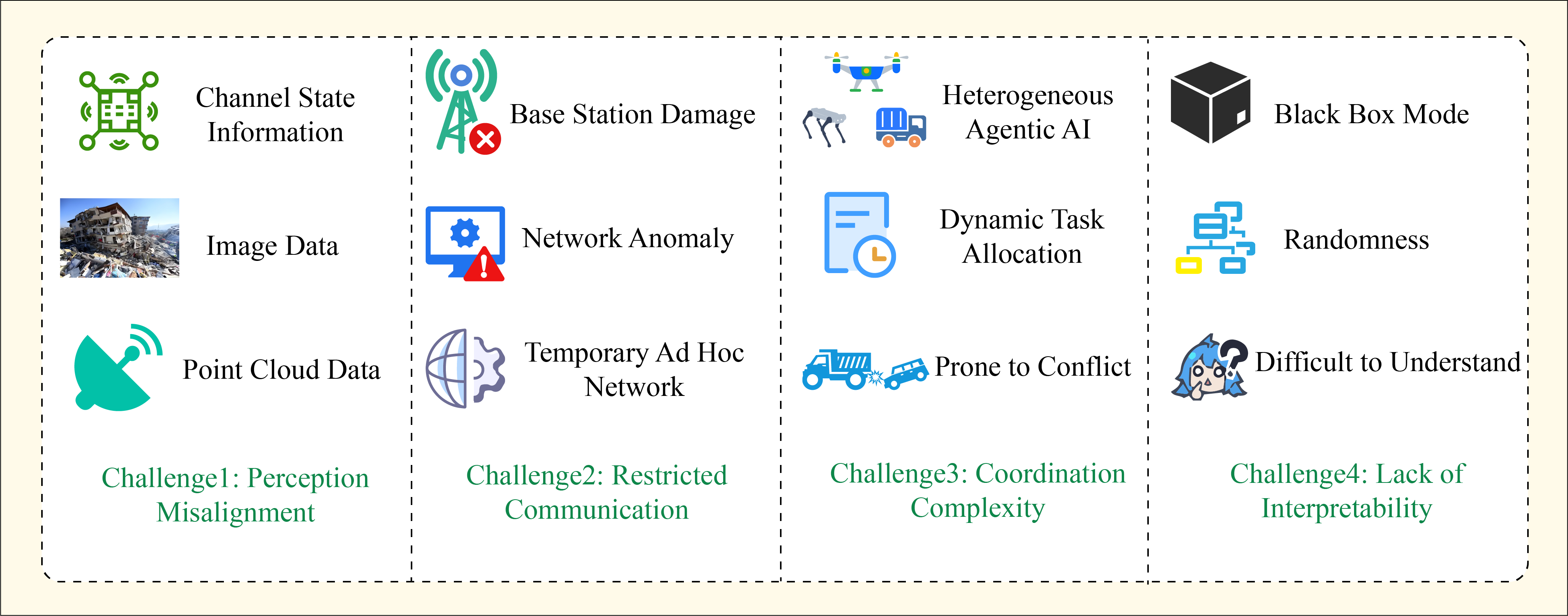}
\caption{Key Challenges in Embodied Intelligence Networks.}
\label{fig1}
\end{figure}
Based on the previous discussion, there are still several challenges to resolve in the EINs. As shown in Fig.~\ref{fig1}, we categorize four key challenges: multimodal semantic information processing, communication and transmission, collaboration among heterogeneous devices, and decision interpretability, elaborated as follows.
\begin{figure*}[htbp]
\centering
\includegraphics[scale=0.4]{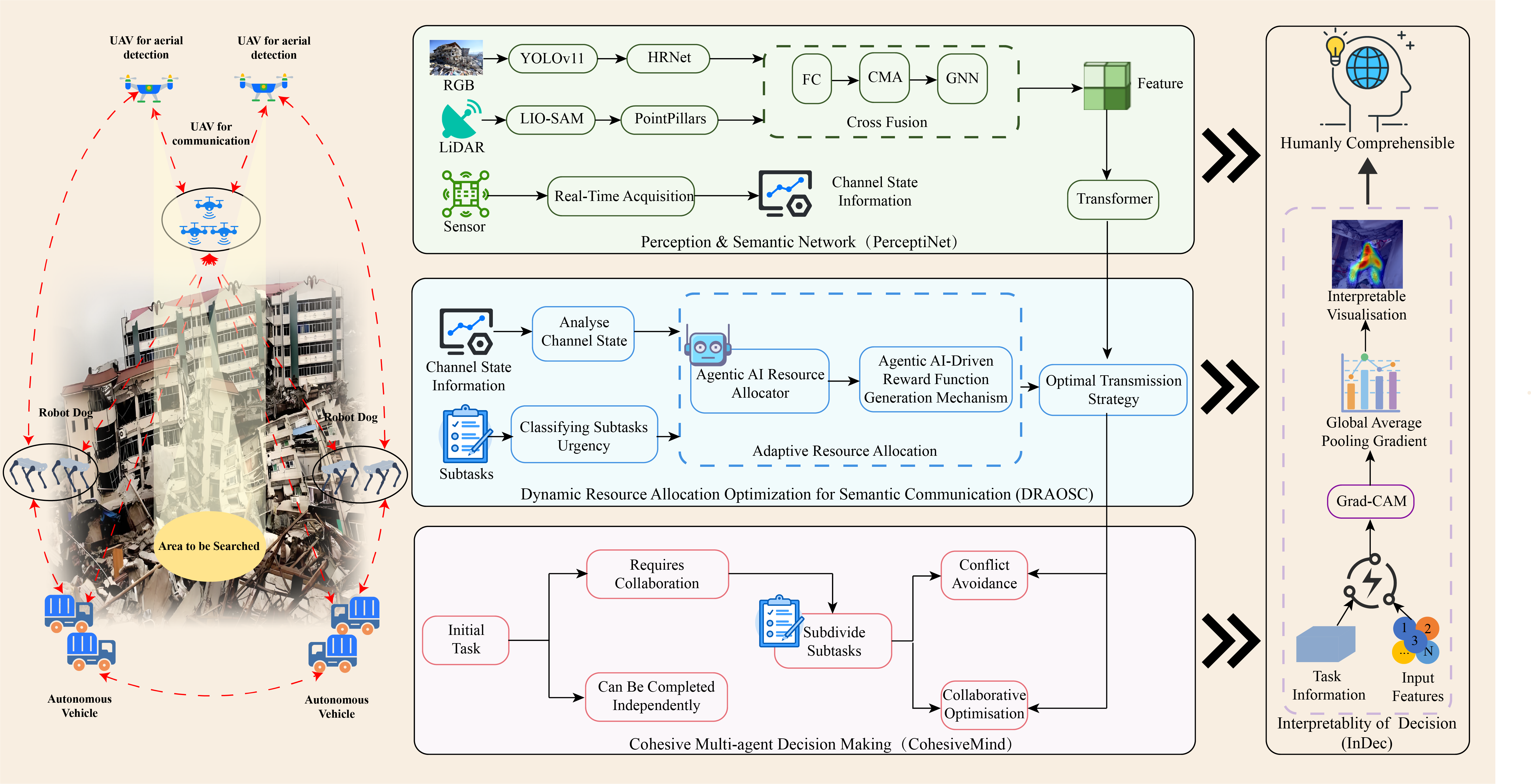}
\caption{The Overall Architecture of CC-EIN Showing PerceptiNet, DRAOSC, CohesiveMind, and InDec.}
\label{fig2}
\end{figure*}
\subsection{Motivation}
\subsubsection{Perception Misalignment}
In an EIN, different types of device typically handle different detection tasks, and the sensory data they collect can vary significantly in type and resolution. Single-modal data has limitations and often cannot fully capture the state and dynamic changes of environmental targets. For instance, visual images provide macro information but are limited under occlusion or poor lighting conditions, while LiDAR point clouds excel in geometric modeling but lack semantic understanding. The key to achieving multimodal semantic communication lies in the efficient alignment and fusion of these multimodal data in semantics, extracting high-level semantic representations that are consistent, compact, and relevant for the task. Moreover, the challenge involves not only feature fusion, but also maintaining semantic consistency between modalities in dynamic environments, forming global situational awareness in embodied intelligent device networks.

\subsubsection{Restricted Communication}
Self-organizing networks or temporary relay among EIDs often encounter issues such as limited bandwidth, uncontrollable latency, and interruption. Relying solely on static configurations or fixed priority strategies will prevent semantic communication from ensuring timely transmission of critical information. Thus, EIDs possess "inherent intelligence." This allows them to dynamically adjust data flows, select optimal transmission paths, and modify data formats. These adjustments are based on task priorities and channel conditions, maximizing information utilization under constrained resources. When the environment changes, the network topology and transmission should be adaptively changed to ensure overall efficiency. In other words, the core of the challenge is integrating adaptive optimization mechanisms into semantic communication, which allows the network to evolve facing both the dynamic environment and the diverse tasks.

\subsubsection{Coordination Complexity}
In a EIN, devices differ significantly in mobility, capacity, sensing, and safety. Its heterogeneity complicates task collaboration because tasks are typically composed of multiple interdependent sub-goals. These sub-goals require different devices to leverage their unique advantages. Moreover, environmental and task conditions can change anytime. Without dynamic task decomposition and real-time adjustments, issues like repeated work, resource conflicts, or task omissions may occur. The main challenge in collaboration is how to effectively decompose tasks based on semantic information, ensuring that each task meets the appropriate device, while enabling conflict detection and rapid resolution. Additionally, the system should support online re-planning and asynchronous feedback mechanisms to address unforeseen situations. It should also find a balance between centralized and distributed scheduling. This ensures that collaboration remains stable and efficient, even with task scaling or local failures.

\subsubsection{Lack of Interpretability}
In uncertain tasks, the transparency of the EIN directly influences human trust in their decisions. Although current Agentic AI models excel in perception and decision-making, their reasoning processes are often a "black-box," lacking intuitive explanatory mechanisms. It does not only diminish human-machine collaboration efficiency, it can also affect the timeliness of decisions during critical moments. To improve the trustworthiness of EID, an interpretability mechanism should integrate into the EIN, allowing one to clearly present their decision through visualization or symbolic methods. The approach improves transparency and accountability while maintaining the performance of the network.

Our article aims to establish a novel collaboration framework that combines semantic communication with MEIDs, enabling "smart communication and intelligent collaboration" throughout the networking process. First, to address the challenge of low correlation between multimodal information, we propose a cross-modal deep fusion approach. This approach extracts key semantic features from different modalities, thus providing a more comprehensive understanding of the environment. Second, to overcome the challenges posed by harsh communication environments and limited bandwidth resources, our goal is to develop an adaptive communication optimization mechanism. This mechanism allows devices to flexibly adjust their communication strategies based on task requirements and network conditions, ensuring reliable transmission of critical semantic information while minimizing bandwidth usage and energy consumption. Third, to tackle task collaboration issues among heterogeneous EIDs, we propose a task decomposition and allocation strategy based on shared semantic information. This strategy ensures efficient cooperation and prevents task overlap or resource contention, thus optimizing overall system performance. Finally, to enhance system credibility and decision transparency, we integrate interpretability into the EIN. Using visualization and symbolic methods, we present the rationale behind decision-making processes, thereby increasing human operators' trust in the devices and improving human-machine collaboration efficiency.

\section{A Case Study: Collaborative Conversational Embodied Intelligence Network}

In this article, faced with the four key challenges, we present CC-EIN, as illustrated in Fig.~\ref{fig2}. The framework comprises four primary functions: PerceptiNet, DRAOSC, CohesiveMind, and InDec. Each specific function works in close collaboration. PerceptiNet extracts high-level semantic information from multimodal data. DRAOSC ensures the efficient transmission of semantic information in complex environments. CohesiveMind formulates the task plans of multiple EIDs based on global semantic information. Finally, InDec provides visual explanations for decision making in the EIDs, helping users to supervise the intelligent devices.

To further validate the effectiveness of the CC-EIN framework, our article proposes a post-disaster rescue scenarios as a case study, illustrating how the framework tackles the collaboration faced by MEIDs in highly uncertain and complex environments.

\subsection{PerceptiNet}
PerceptiNet is responsible for extracting unified semantic representations from the multimodal data in each embodied intelligent device. In this framework, drones, autonomous vehicles, and robot dogs are equipped with cameras and LiDAR, while they also integrate the evaluation of environment to detect the channel state in real time. Then it provides the multimodal environment data for DRAOSC. The entire perception is powered by Agentic AI, which autonomously extracts, integrates, and encodes multimodal information, offering strong support for DRAOSC and CohesiveMind.

During the local perception stage, each embodied intelligent device processes the collected data using its internal Agentic AI. Specifically, the drone employs a visual agent with YOLOv11 and HRNet to extract key visual features, the autonomous vehicle utilizes a radar point cloud agent with LIO-SAM and PointPillars to handle LiDAR point cloud data, and the robot dog integrates data from a close-range camera and point cloud agent to compensate for macro-level perception blind spots. Meanwhile, communication environment data are analyzed and encoded by a dedicated communication agent, providing intelligent input for adaptive transmission optimization and thereby enabling efficient coordination between perception and communication.

\subsection{DRAOSC}
\begin{figure}[htbp]
\centering
\includegraphics[scale=0.23]{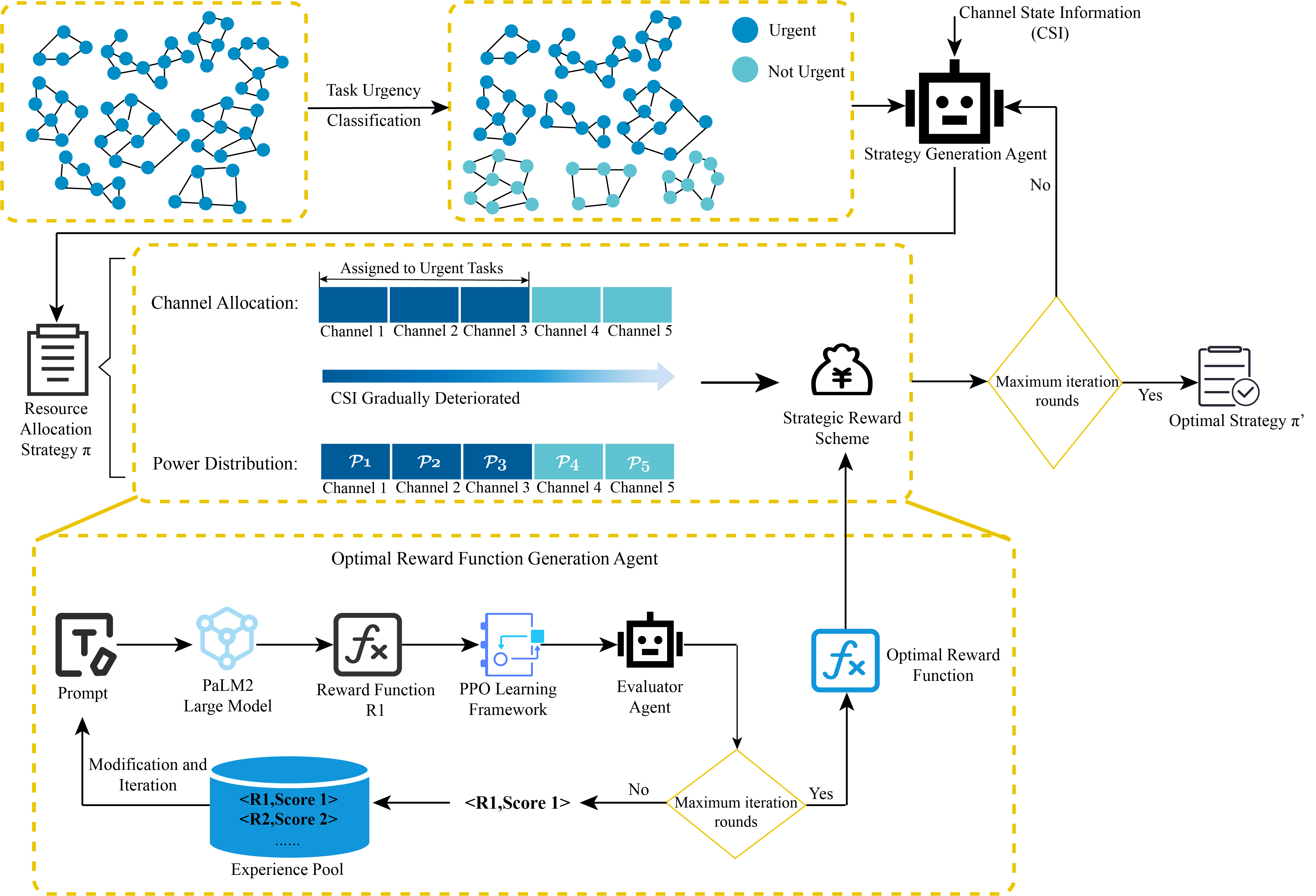}
\caption{The Adaptive Transmission Optimization Process of DRAOSC in CC-EIN.}
\label{fig3}
\end{figure}

The DRAOSC workflow is shown in Fig.~\ref{fig3}. The DRAOSC module is designed to ensure efficient and reliable transmission of semantic information under bandwidth constraints and unstable channel conditions, with the entire process driven by Agentic AI. By autonomously sensing network states and dynamically adjusting communication parameters according to task priorities, Agentic AI guarantees efficient delivery of critical semantic data. Specifically, DRAOSC classifies subtasks by urgency and continuously monitors channel conditions, such as signal-to-noise ratio and bandwidth utilization, to provide intelligent input for resource allocation.

The communication optimization problem is formulated as a constrained decision-making task and is addressed collaboratively by multiple agents. The policy generation agent derives optimal strategies from the state space (task urgency and channel conditions) and the action space (channel selection and power allocation). The reward function generation agent incorporates metrics such as transmission success rate, latency, packet loss rate, bandwidth, and energy consumption, and leverages the large language model PaLM2 to automatically construct and refine the reward function. The evaluator agent then assesses the performance of the policies during training, assigns scores, and allows closed-loop optimization. The Proximal Policy Optimization (PPO) algorithm is employed to train these agents, allowing them to learn optimal strategies in continuous state–action spaces with improved stability.

Once training is completed, DRAOSC adaptively adjusts its strategies according to the prevailing network conditions. Under favorable conditions, EID can transmit multiple streams of information in parallel, whereas in degraded channels the system increases transmission power for urgent data and defers secondary information until the channel recovers.

\subsection{CohesiveMind}
CohesiveMind functions as the central brain of the embodied intelligent device system, powered by internal Agentic AI to plan tasks and coordinate device collaboration based on global semantic information. Upon receiving high-level task commands, the task parsing agent interprets the tasks with reference to the environmental semantic model provided by PerceptiNet to accurately identify task requirements. For tasks that cannot be accomplished by a single device or are suitable for parallel execution, the task decomposition agent divides them into semantic collaboration instructions and disseminates these via semantic communication to the appropriate devices, ensuring clear allocation and well-defined objectives.

During execution, devices employ the strategy adjustment agent to interpret instructions and formulate execution plans. CohesiveMind continuously monitors task progress, and when devices face constraints or task requirements evolve, the strategy adjustment agent dynamically reallocates assignments to maintain smooth execution. In parallel, the system integrates conflict avoidance and optimization mechanisms, where the strategy adjustment agent adjusts resource allocation according to device capabilities and task demands. This design enables EIDs to collaborate efficiently in complex tasks, minimizing conflicts and redundant operations while enhancing the reliability and efficiency of task completion.
\subsection{InDec}
To enhance the transparency and reliability of decision-making in MEIDs systems, we introduce a dedicated InDec module, which analyzes the key factors behind device decisions and presents them in a human-understandable format. In particular, we apply Grad-CAM-based deep visual explanation techniques to generate heatmaps that indicate the regions of EIDs focus when the devices process images. This allows operators to interpret the rationale behind each decision through visualization, thereby improving understanding of the devices’ behavior.
\begin{figure}[htbp]
\centering
\includegraphics[scale=0.38]{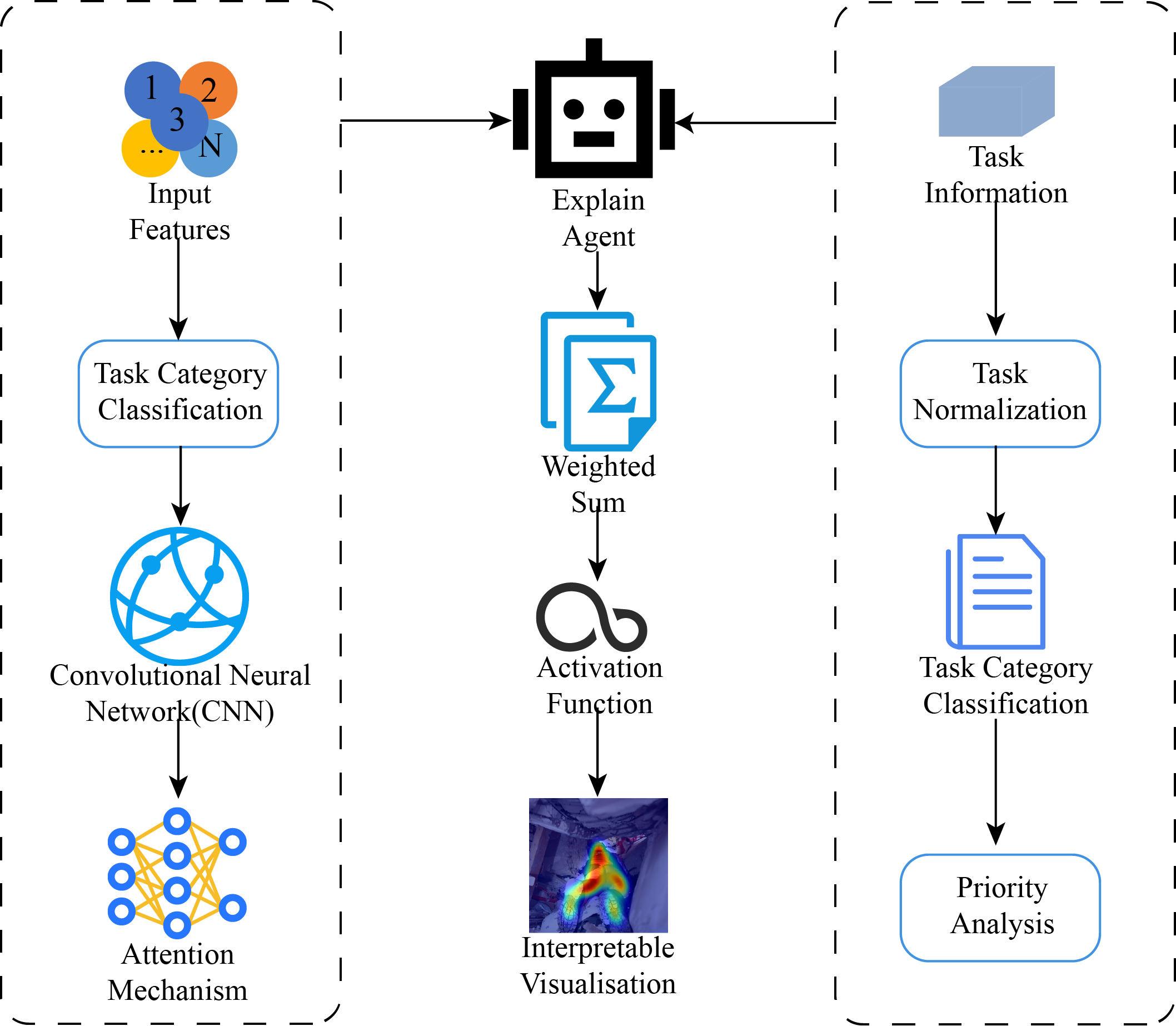}
\caption{The Grad-CAM Visualization Mechanism of InDec.}
\label{fig4}
\end{figure}

The InDec workflow is shown in Fig.~\ref{fig4}, where its Interpretable visualization links the input features with the decisions of the EIDs. In this process, a convolutional neural network (CNN) first extracts input feature maps, which are then analyzed by a dedicated explain agent to compute the partial derivatives of the decision output with respect to each convolutional feature channel. The resulting channel weights are applied to the feature maps to generate activation maps, filtered through ReLU to remove negative values, and finally overlaid on the original image to highlight key regions that support the decision.

By using InDec, the system can clearly demonstrate the relationship between the decision-making process and input features, mitigating the “black-box” issue and enhancing transparency and accountability. Furthermore, this Interpretable information can be fed back to DRAOSC and CohesiveMind to optimize communication strategies and task coordination. When InDec highlights critical features that heavily influence a decision, DRAOSC can prioritize the transmission of relevant semantic information, while CohesiveMind can adjust task allocation or collaboration sequences accordingly, thereby improving overall system efficiency and human–machine collaboration.

\section{Performance Evaluation}
To validate the framework’s effectiveness, we constructed a post-disaster urban simulation environment including collapsed buildings, road obstacles, and trapped individuals. Rescue tasks are carried out by drones, autonomous vehicles, and robot dogs, responsible for large-area target search, path search and supply delivery, and close-range search and annotation within collapsed buildings, respectively. All devices communicate via a self-organizing network, simulating a damaged 4G cellular area with bandwidth fluctuations from 500 MHz down to 50 MHz to evaluate communication system adaptability. Four metrics are used for evaluation: TCR, measuring overall multi-agent collaboration effectiveness; TE, defined as 'number of completed tasks / transmitted data (MB)', reflecting the ratio between communication load and task benefit; average transmission power under varying bandwidths, recorded as bandwidth decreases from 500 MHz to 50 MHz to assess the communication module’s adaptability under constraints; and SC, indicating the agreement of EIDs’ semantic understanding for task-relevant information. Standard semantic descriptions from the knowledge base serve as references to compare EIDs’ detection results and attribute annotations for key targets, producing a consistency score between 0 and 1, with 1 representing perfect consistency. We evaluate four schemes in the same environment: (1) the complete CC-EIN framework; (2) CC-EIN (w/o DRAOSC); (3) the GA-PPO method \cite{14}; and (4) the CF method \cite{15}.
\begin{figure}[htbp]
\centering
\includegraphics[scale=0.32]{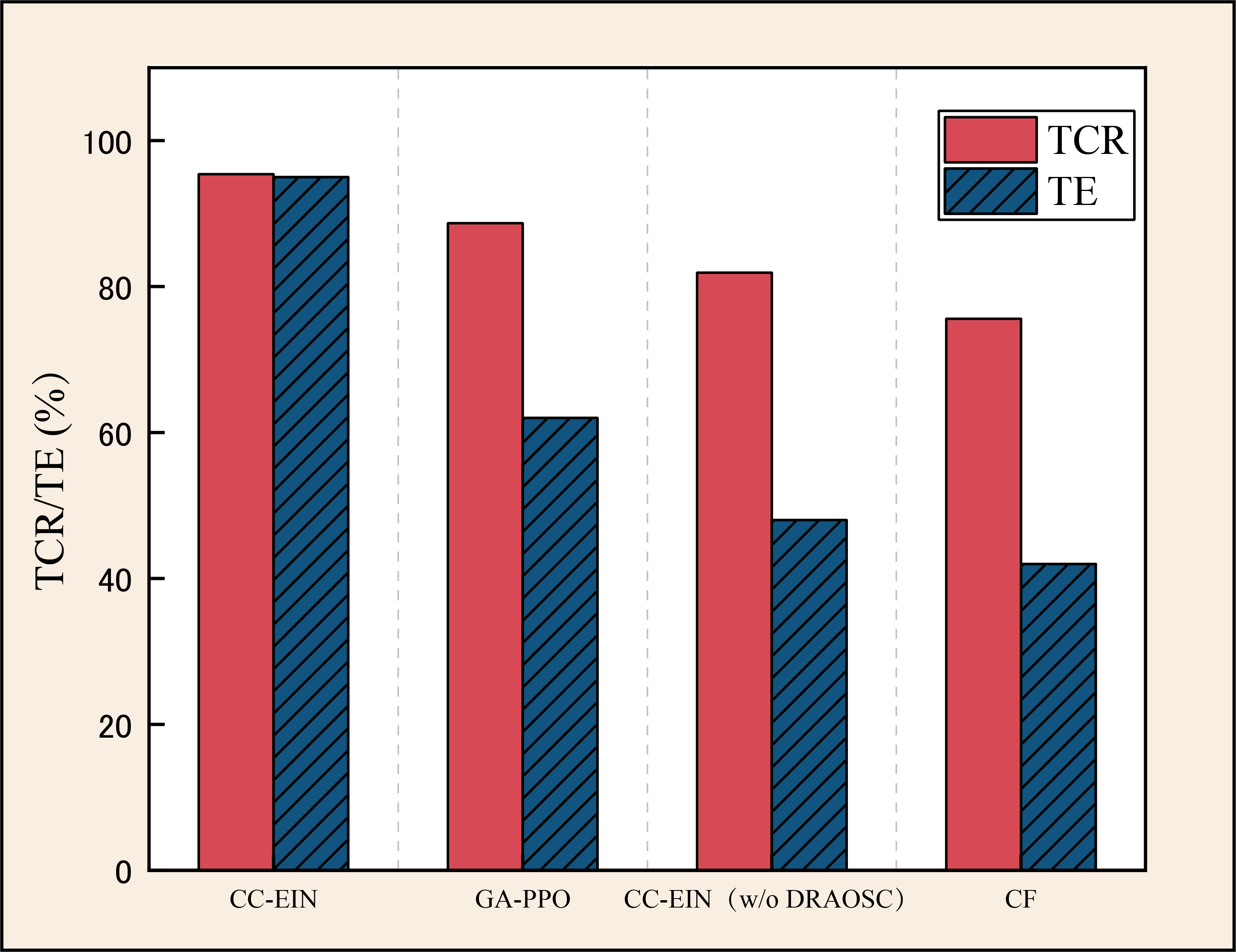}
\caption{Comparison of TCR and TE Across Different Methods, Demonstrating the Collaboration Performance and Resource Utilization of MEIDs.}
\label{fig5}
\end{figure}

Fig.~\ref{fig5} compares TCR and TE, clearly demonstrating that the CC-EIN framework outperforms the other three methods. CC-EIN achieves a TCR of 95.4\%, significantly higher than GA-PPO (88.7\%), CC-EIN without DRAOSC (81.9\%), and CF (75.6\%), exceeding them by 6.7\%, 13.5\%, and 19.8\%, respectively. In TE, CC-EIN leads with 95\%, far surpassing GA-PPO (62\%), CC-EIN without DRAOSC (48\%), and CF (42\%), with improvements of 33\%, 47\%, and 53\%, respectively. These results indicate that CC-EIN not only has a clear advantage in task completion but also performs better in resource utilization and collaborative efficiency.

\begin{figure}[htbp]
\centering
\includegraphics[scale=0.32]{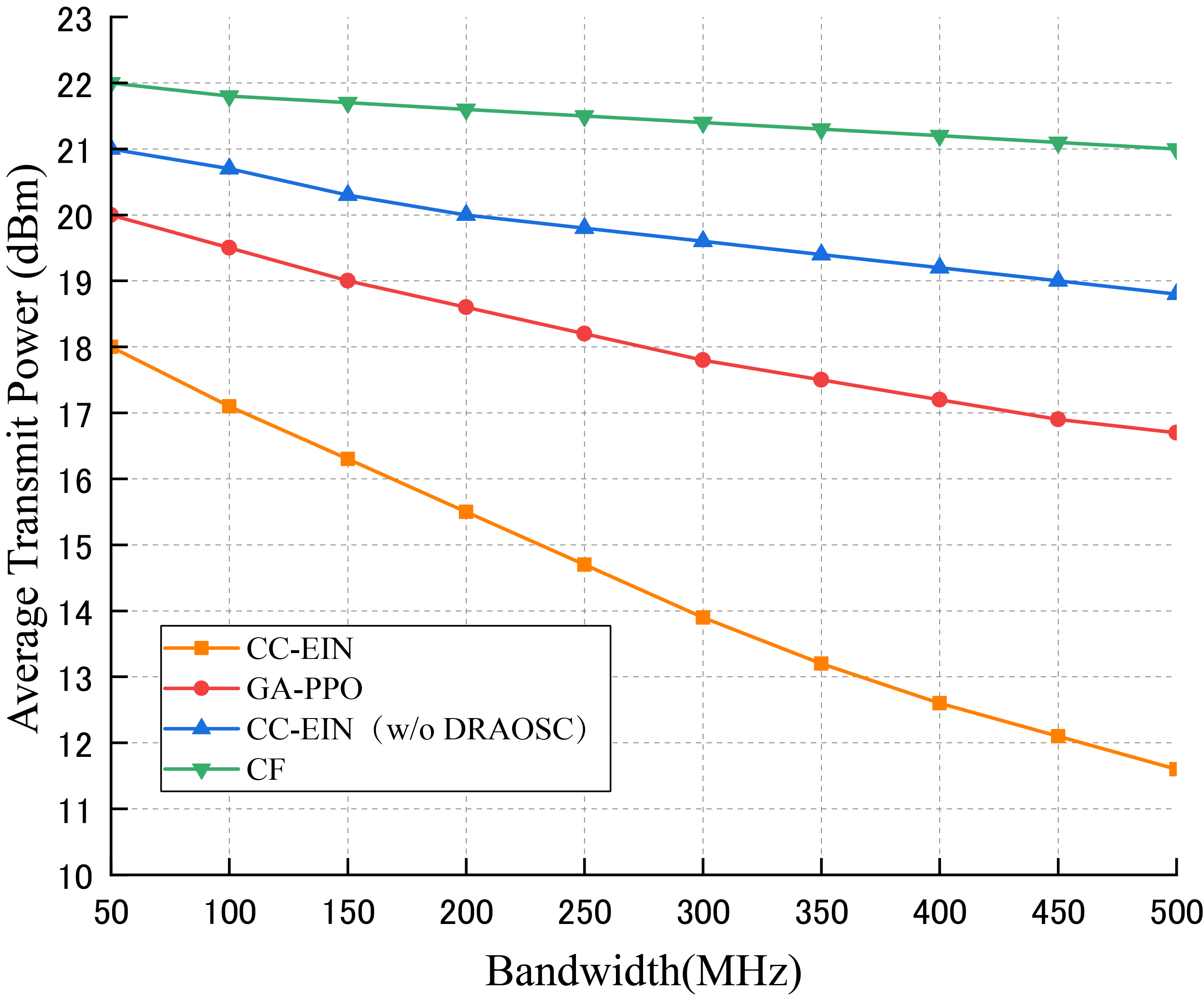}
\caption{Average Transmission Power under Varying Bandwidths, Illustrating Adaptive Power Control and Communication Efficiency Across Methods.}
\label{fig6}
\end{figure}
Fig.~\ref{fig6} shows power consumption under different bandwidths. Experiments, conducted with consistent network conditions, channel states, task types, and device configurations, record the power usage of each method across varying bandwidths. CC-EIN consistently consumes significantly less power than the other methods from 50 MHz to 500 MHz and effectively adjusts its power usage according to bandwidth, maintaining stable communication efficiency. At 50 MHz, CC-EIN’s transmission power is 18 dBm, lower than GA-PPO (20 dBm), CC-EIN without DRAOSC (21 dBm), and CF (22 dBm). As bandwidth increases, CC-EIN’s power steadily decreases to 11.6 dBm at 500 MHz, further demonstrating its efficiency and energy-saving advantage at high bandwidths. In contrast, other methods show higher power consumption with less variation, indicating that CC-EIN maintains good resource utilization as bandwidth grows. These results also highlight the key role of the DRAOSC module in enhancing communication efficiency and reducing power usage, ensuring efficient task execution even under bandwidth-limited conditions.

\begin{figure}[htbp]
\centering
\includegraphics[scale=0.32]{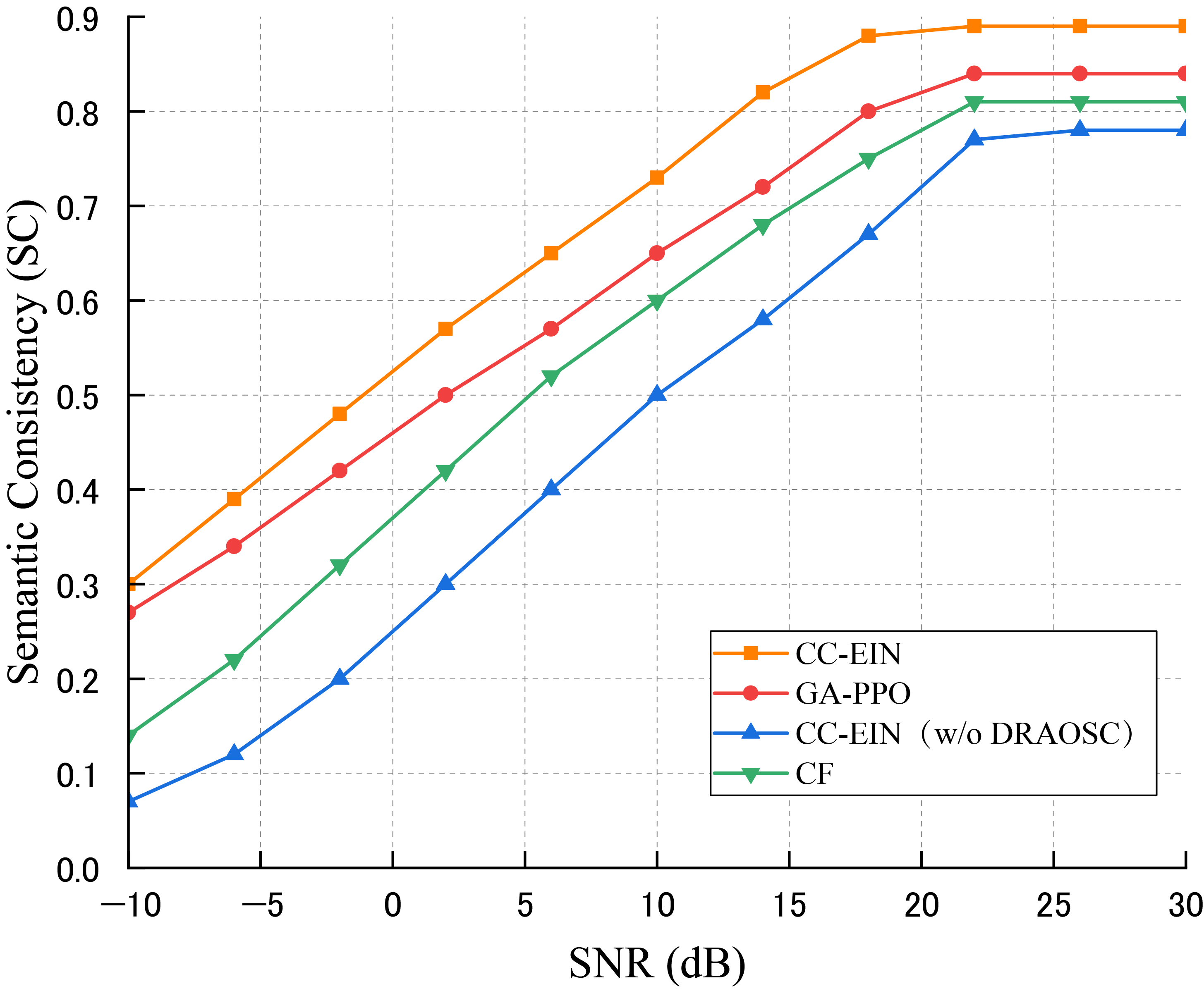}
\caption{SC of MEIDs under Varying SNRs, Demonstrating the Accuracy of Task-Related Information Transmission Across Methods.}
\label{fig7}
\end{figure}

Fig.~\ref{fig7} shows SC of each method under varying signal-to-noise ratios (SNRs). Experimental results indicate that CC-EIN maintains high SC across all conditions. At higher SNRs, its SC approaches 1, highlighting its accuracy and reliability in task information transmission. At 30 dB, CC-EIN achieves an SC of 0.89, significantly outperforming GA-PPO (0.84), CC-EIN without DRAOSC (0.78), and CF (0.81), demonstrating consistent performance under different SNRs. Even at low SNRs (e.g., -10 dB), CC-EIN remains stable with an SC of 0.3, compared to GA-PPO (0.27), CC-EIN without DRAOSC (0.07), and CF (0.14). These results confirm CC-EIN’s strong advantage in preserving semantic consistency across various SNR conditions, especially in complex environments, thereby improving MEIDs' collaboration efficiency.

\begin{figure}[htbp]
\centering
\includegraphics[scale=0.3]{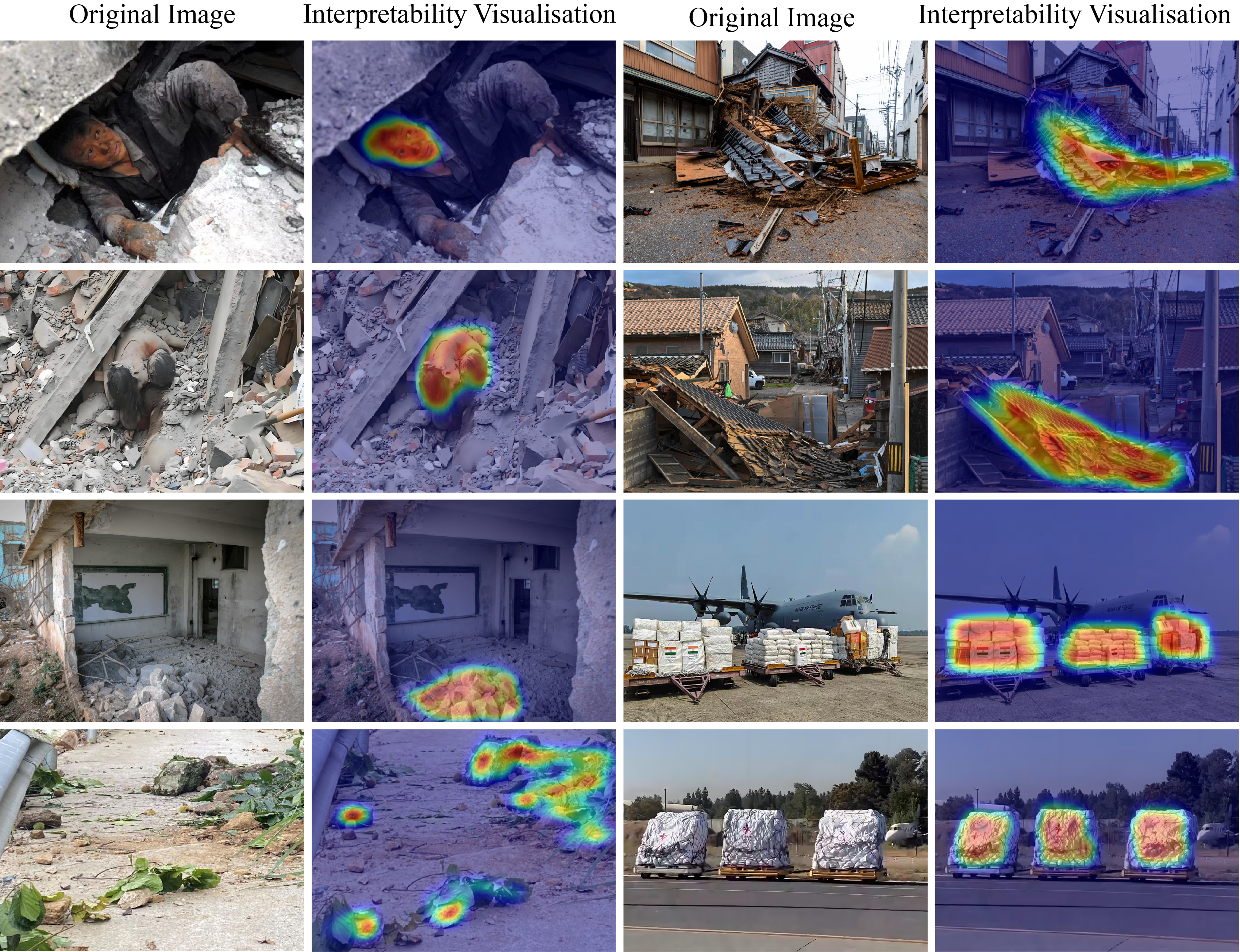}
\caption{Interpretable Visualizations of the InDec Module Across Tasks.}
\label{fig8}
\end{figure}

Fig.~\ref{fig8} presents the interpretable visualizations of the InDec module on different tasks. Grad-CAM highlights the areas of interest of EIDs during decision making. In the victim rescue task, the heatmap focuses on the victims’ locations, reflecting decisions based on visual input. In the road detour task, it emphasizes obstacles, indicating the rationale for path planning and obstacle avoidance. In the road clearance and supply delivery tasks, the heatmaps show attention on obstacles and supplies. These visualizations enhance decision-making transparency and improve human–machine collaboration efficiency and trust.

\section{Conclusion}
This article presents a CC-EIN for MEIDs, structured around four functional components: PerceptiNet, DRAOSC, CohesiveMind, and InDec. First, PerceptiNet collects visual images, radar signals, and environmental parameters, aligning them into a unified semantic representation that supports the understanding of high-level tasks in complex rescue scenarios. Second, DRAOSC dynamically adjusts coding schemes, compression ratios, and transmission power according to task urgency and channel conditions, thus improving spectral efficiency and ensuring the reliable delivery of critical semantics. Third, CohesiveMind utilizes a shared semantic knowledge base to decompose rescue tasks, match them to appropriate agents such as drones, autonomous vehicles, or robot dogs, and resolve potential conflicts to ensure coordinated execution. Finally, InDec applies Grad-CAM to visualize decision rationales highlighting factors such as survivor detection, obstacle removal, or path re-planning, thus providing transparent insights into agent behaviors and improving human–AI trust in time-critical operations. Interpretable decision-making uses Grad-CAM to highlight attention areas and semantic contributions, enhance transparency, and support human intervention.
In post-disaster rescue simulations, the framework achieves 95.4\% TCR and 95\% TE, while adaptively lowering average transmission power as bandwidth increases, demonstrating strong energy efficiency and communication reliability. Overall, the framework provides a practical, interpretable system for 6G-enabled multi-agent collaboration. Future work will focus on improving multimodal fusion, adaptive communication, and task scheduling under dynamic conditions, with interpretable decision making remaining key to reliable collaboration.
\bibliographystyle{IEEEtran}
\bibliography{reference}

@ARTICLE{1,
  author={Lin, Zheng and Qu, Guanqiao and Chen, Qiyuan and Chen, Xianhao and Chen, Zhe and Huang, Kaibin},
  journal={IEEE Communications Magazine}, 
  title={Pushing Large Language Models to the 6G Edge: Vision, Challenges, and Opportunities}, 
  year={2025},
  volume={63},
  number={9},
  pages={52-59},
  doi={10.1109/MCOM.001.2400764}}

@ARTICLE{2,
  author={Zhang, Yuexia and Wang, Xinyi and Gang, Yuanshuo and Wang, Jian and Wu, Sheng and Zhang, Peiying and Shi, Yuanming},
  journal={IEEE Communications Magazine}, 
  title={6G SAGIN Information Transmission Model}, 
  year={2025},
  volume={63},
  number={6},
  pages={98-105},
  doi={10.1109/MCOM.001.2400351}}

@ARTICLE{3,
  author={Rodríguez-Piñeiro, José and Wei, Zhongxiang and Wang, Jingjing and Gutiérrez, Carlos A. and Correia, Luis M.},
  journal={IEEE Open Journal of Vehicular Technology}, 
  title={6G-Enabled Vehicle-to-Everything Communications: Current Research Trends and Open Challenges}, 
  year={2025},
  volume={6},
  number={},
  pages={2358-2391},
  doi={10.1109/OJVT.2025.3599570}}

@ARTICLE{4,
  author={Tan, Jinbiao and Shi, Jianhua and Wu, Ligang and Chen, Baotong and Tang, Hao and Zhang, Chunhua and Zhang, Wujie and Wang, Shiyong and Wan, Jiafu},
  journal={IEEE Access}, 
  title={Embodied Intelligence Empowering Customized Manufacturing: Architecture, Opportunities, and Challenges}, 
  year={2025},
  volume={13},
  number={},
  pages={92740-92755},
  doi={10.1109/ACCESS.2025.3572778}}

@ARTICLE{5,
  author={Acharya, Deepak Bhaskar and Kuppan, Karthigeyan and Divya, B.},
  journal={IEEE Access}, 
  title={Agentic AI: Autonomous Intelligence for Complex Goals—A Comprehensive Survey}, 
  year={2025},
  volume={13},
  number={},
  pages={18912-18936},
  doi={10.1109/ACCESS.2025.3532853}}

@ARTICLE{6,
  author={Ren, Lei and Dong, Jiabao and Liu, Shuai and Zhang, Lin and Wang, Lihui},
  journal={IEEE/ASME Transactions on Mechatronics}, 
  title={Embodied Intelligence Toward Future Smart Manufacturing in the Era of AI Foundation Model}, 
  year={2025},
  volume={30},
  number={4},
  pages={2632-2642},
  doi={10.1109/TMECH.2024.3456250}}

@ARTICLE{7,
  author={Shen, Tianyu and Sun, Jinlin and Kong, Shihan and Wang, Yutong and Li, Juanjuan and Li, Xuan and Wang, Fei-Yue},
  journal={IEEE/CAA Journal of Automatica Sinica}, 
  title={The Journey/DAO/TAO of Embodied Intelligence: From Large Models to Foundation Intelligence and Parallel Intelligence}, 
  year={2024},
  volume={11},
  number={6},
  pages={1313-1316},
  doi={10.1109/JAS.2024.124407}}

@ARTICLE{8,
  author={Wu, Hefeng and Jiang, Hao and Wang, Keze and Tang, Ziyi and He, Xianghuan and Lin, Liang},
  journal={IEEE Transactions on Multimedia}, 
  title={Improving Network Interpretability via Explanation Consistency Evaluation}, 
  year={2024},
  volume={26},
  number={},
  pages={11261-11273},
  doi={10.1109/TMM.2024.3453058}}

@ARTICLE{9,
  author={Gan, Yahui and Zhang, Bo and Shao, Jiawei and Han, Zao and Li, Ang and Dai, Xianzhong},
  journal={IEEE Robotics and Automation Letters}, 
  title={Embodied Intelligence: Bionic Robot Controller Integrating Environment Perception, Autonomous Planning, and Motion Control}, 
  year={2024},
  volume={9},
  number={5},
  pages={4559-4566},
  doi={10.1109/LRA.2024.3377559}}

@ARTICLE{10,
  author={Jin, Zhu and Song, Tiecheng and Jia, Wen-Kang and Zou, Wenbin and Song, Xiaoqin},
  journal={IEEE Internet of Things Journal}, 
  title={Task-Oriented Semantic Communication With Adaptive Semantic Reconstruction Network}, 
  year={2025},
  volume={12},
  number={17},
  pages={35784-35798},
  doi={10.1109/JIOT.2025.3579582}}

@ARTICLE{11,
  author={Ding, Guangyao and Liu, Shengli and Yuan, Jiantao and Yu, Guanding},
  journal={IEEE Transactions on Wireless Communications}, 
  title={Joint URLLC Traffic Scheduling and Resource Allocation for Semantic Communication Systems}, 
  year={2024},
  volume={23},
  number={7},
  pages={7278-7290},
  doi={10.1109/TWC.2023.3339239}}

@ARTICLE{12,
  author={Sun, Haofeng and Ni, Wanli and Tian, Hui and Zheng, Jingheng and Nie, Gaofeng and Niyato, Dusit},
  journal={IEEE Internet of Things Journal}, 
  title={A Hybrid Federated Learning Framework for Task-Oriented Semantic Communication}, 
  year={2025},
  volume={12},
  number={13},
  pages={23444-23461},
  doi={10.1109/JIOT.2025.3553504}}

@ARTICLE{13,
  author={Hu, Kai and Zou, Longhao and Chen, Zihao and Jiang, Jun and Jiang, Fan and Tao, Xiaofeng},
  journal={IEEE Network}, 
  title={Wireless Multi-Robot Collaboration: Communications, Perception, Control, and Planning}, 
  year={2025},
  volume={39},
  number={4},
  pages={302-311},
  doi={10.1109/MNET.2024.3483829}}

@Article{14,
AUTHOR = {Fang, Zheng and Ma, Tao and Huang, Jun and Niu, Zhao and Yang, Fang},
JOURNAL = {Applied Sciences},
TITLE = {Efficient Task Allocation in Multi-Agent Systems Using Reinforcement Learning and Genetic Algorithm},
YEAR = {2025},
VOLUME = {15},
NUMBER = {4},
ARTICLE-NUMBER = {1905},
DOI = {10.3390/app15041905}}

@Article{15,
AUTHOR = {Liu, Lixiang and Li, Peng},
JOURNAL = {Vehicles},
TITLE = {Game-Theoretic Cooperative Task Allocation for Multiple-Mobile-Robot Systems},
YEAR = {2025},
VOLUME = {7},
NUMBER = {2},
ARTICLE-NUMBER = {35},
DOI = {10.3390/vehicles7020035}}



%








\end{document}